\documentclass[sigconf]{acmart}

\AtBeginDocument{%
  \providecommand\BibTeX{{%
    \normalfont B\kern-0.5em{\scshape i\kern-0.25em b}\kern-0.8em\TeX}}}

\setcopyright{none}
\copyrightyear{2024}
\acmYear{2024}
\acmDOI{}
\acmPrice{}
\acmISBN{}
\renewcommand\footnotetextcopyrightpermission[1]{} 

\usepackage{amsmath,amsfonts,bm}

\def\eqref#1{equation~\ref{#1}}

\def\1{\bm{1}}

\DeclareMathAlphabet{\mathsfit}{\encodingdefault}{\sfdefault}{m}{sl}
\SetMathAlphabet{\mathsfit}{bold}{\encodingdefault}{\sfdefault}{bx}{n}

\newcommand{\softmax}{\mathrm{softmax}}

\usepackage[most]{tcolorbox}
\usepackage{caption}
\usepackage{multirow}
\usepackage{enumitem}
\usepackage{subfig}
\usepackage{color, colortbl}
\usepackage{makecell}
\usepackage[symbol]{footmisc}

\definecolor{Gray}{gray}{0.9}
\begin{document}

\title{Diet-ODIN: A Novel Framework for Opioid Misuse Detection with Interpretable Dietary Patterns}

\author{Zheyuan Zhang*, Zehong Wang*, Shifu Hou}
\thanks{*Both authors contributed equally. \quad\quad\quad $\dagger$ Corresponding author} 
\affiliation{%
 \institution{University of Notre Dame}
 \city{Notre Dame}
 \state{Indiana}
 \country{USA}}
\email{{zzhang42, zwang43, shou}@nd.edu}

\author{Evan Hall, Landon Bachman, Vincent Galassi}
\affiliation{%
 \institution{University of Notre Dame}
 \city{Notre Dame}
 \state{Indiana}
 \country{USA}}
\email{{ehall9, lbachma2, vgalassi}@nd.edu }

\author{Jasmine White}
\affiliation{%
 \institution{Purdue University}
 \city{West Lafayette}
 \state{Indiana}
 \country{USA}}
\email{white742@purdue.edu }

\author{Nitesh V. Chawla}
\affiliation{%
 \institution{University of Notre Dame}
 \city{Notre Dame}
 \state{Indiana}
 \country{USA}}
\email{nchawla@nd.edu}

\author{Chuxu Zhang}
\affiliation{%
 \institution{Brandeis University}
 \city{Waltham}
 \state{Massachusetts}
 \country{USA}}
\email{chuxuzhang@brandeis.edu}

\author{Yanfang Ye$^\dagger$} 
\affiliation{%
 \institution{University of Notre Dame}
 \city{Notre Dame}
 \state{Indiana}
 \country{USA}}
\email{yye7@nd.edu}

\renewcommand{\shortauthors}{Zhang and Wang, et al.}

\begin{abstract}

    The opioid crisis has been one of the most critical society concerns in the United States. Although the medication assisted treatment (MAT) is recognized as the most effective treatment for opioid misuse and addiction, the various side effects can trigger opioid relapse. In addition to MAT, the dietary nutrition intervention has been demonstrated its importance in opioid misuse prevention and recovery. However, research on the alarming connections between dietary patterns and opioid misuse remain under-explored. In response to this gap, in this paper, we first establish a large-scale multifaceted dietary benchmark dataset related to opioid users \textit{at the first attempt} and then develop a novel framework - i.e., namely \textbf{O}pioid Misuse \textbf{D}etection with \textbf{IN}terpretable \textbf{Diet}ary Patterns (\textbf{Diet-ODIN}) - to bridge heterogeneous graph (HG) and large language model (LLM) for the identification of users with opioid misuse and the interpretation of their associated dietary patterns. Specifically, in Diet-ODIN, we first construct an HG to comprehensively incorporate both dietary and health-related information, and then we devise a holistic graph learning framework with noise reduction to fully capitalize both users' individual dietary habits and shared dietary patterns for the detection of users with opioid misuse. To further delve into the intricate correlations between dietary patterns and opioid misuse, we exploit an LLM by utilizing the knowledge obtained from the graph learning model for interpretation. The extensive experimental results based on our established benchmark with quantitative and qualitative measures demonstrate the outstanding performance of Diet-ODIN on exploring the complex interplay between opioid misuse and dietary patterns, by comparison with state-of-the-art baseline methods. Our code, built benchmark and system demo are available at \href{https://github.com/JasonZhangzy1757/Diet-ODIN}{https:/github.com/Diet-ODIN}.

\end{abstract}

\maketitle
\pagestyle{plain}

\section{Introduction}

Opioids are a class of drugs including the illegal drug heroin, synthetic opioids like fentanyl, and prescription pain relievers such as oxycodone \cite{NIDA-Opioid}. Besides pain relief, opioids could produce euphoria and therefore might be easily misused \cite{dennett202opdiet,rigg2010motivations,rosenblum2008opioids} - e.g., 10.1 million Americans reported misusing opioids in 2019; an estimated 108,000 drug overdose deaths in the United States in 2021 \cite{CDC-OPmisuse}, 90\% of which involved opioids \cite{CDC-OPdeath}. Besides the negative impact on public health, opioid misuse and addiction have also caused devastating socioeconomic crises (e.g., domestic violence, child abuse). To battle the deadly opioid epidemic, medication assisted treatment (MAT) \cite{center2005medication}, which takes the opioid agonists such as methadone or buprenorphine in combination with counseling, has been recognized as the most effective treatment for opioid misuse and addiction \cite{mcdonald2019hedonic}. Despite its effectiveness, the various induced side-effects can often trigger opioid relapse \cite{dennett202opdiet,mcdonald2019hedonic,nabipour2014burden}, which calls for novel solutions to promote health resilience for people against the opioid epidemic. 

\begin{figure*}[htbp!]
	\centering
	\includegraphics[width=1\linewidth]{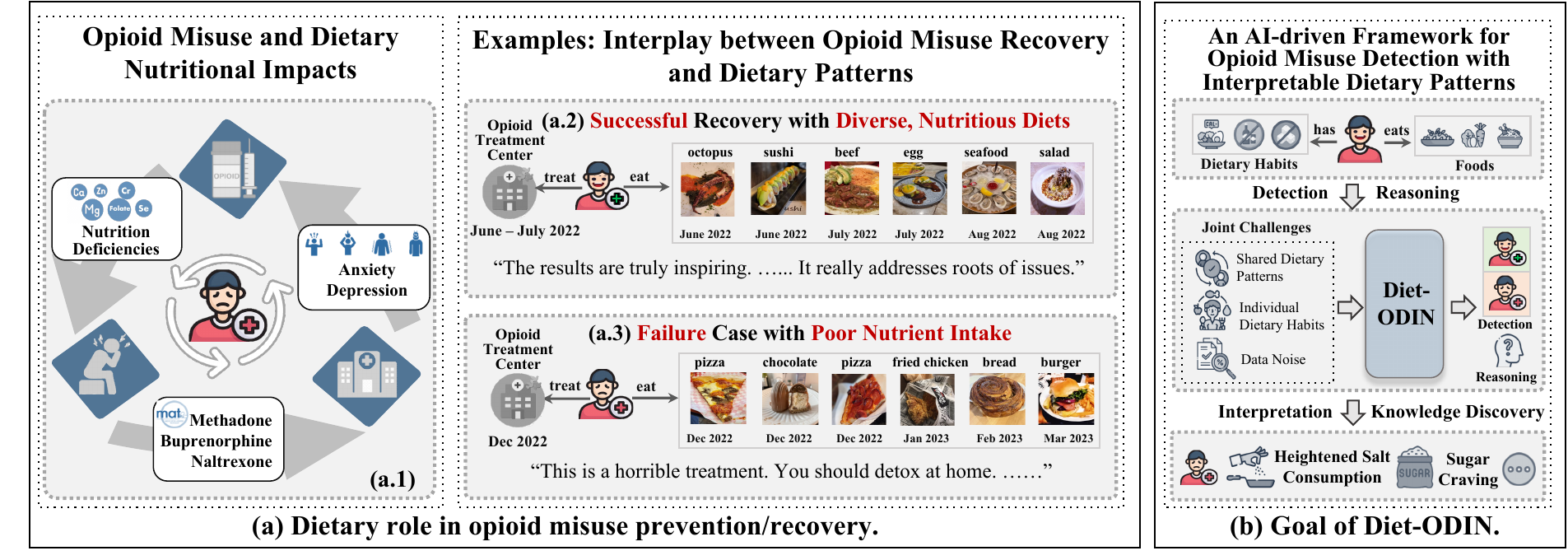}
	\vspace{-0.5cm}
	\caption{ The illustration of (a) dietary role in opioid misuse prevention/recovery, and (b) goal of Diet-ODIN.}
    \vspace{-5pt}
    \label{fig-1:intro}
\end{figure*}

In addition to MAT, it has been demonstrated that dietary nutrition intervention could play an important role in opioid misuse prevention and recovery \cite{dennett202opdiet,chavez2020nutritional,kheradmand2020nutritional,jeynes2017importance,waddington2015nutritional,nabipour2014burden}. As illustrated in Figure \ref{fig-1:intro}.(a.1), the research has shown that opioid misuse can often lead to malnutrition, nutrient deficiencies such as reduced protein~\cite{mahboub2021nutritional} and dietary fiber intake~\cite{ketwaroo2013opioid}, metabolic disorders that compromise nutrition, and altered body composition, which in turn impede its prevention and recovery; while Figure \ref{fig-1:intro}.(a.2) and (a.3) have also exemplified the interplay between opioid misuse recovery and associated dietary patterns. These indicate that the distinct dietary patterns may serve as subtle yet informative indicators for the early detection of opioid misuse, particularly in scenarios such as self-medication or poly-pharmacy, where individuals might not recognize their misuse~\cite{matos2020opioids}, or in cases where misuse is intentionally concealed. Moreover, the task can not only assist in the timely identification of at-risk individuals for healthcare providers and relevant authorities to intervene, but also pave a new paradigm in opioid misuse and addition treatment through the understanding of dietary patterns linked to opioid misuse. This naturally brings up the question: \textit{how can we leverage dietary data to identify users with potential opioid misuse, and enhance the understanding of both known and less-known dietary patterns associated with opioid misuse?}

To answer this question, existing works \cite{kheradmand2020nutritional,mcdonald2019hedonic,jeynes2017importance,waddington2015nutritional,nabipour2014burden} predominantly rely on traditional methods, such as limited surveys and case studies. There exists a notable gap in automating larger scale data analysis to comprehensively explore the alarming connections between dietary patterns and opioid misuse. To bridge this gap, as shown in Figure \ref{fig-1:intro}.(b), in this work, we establish a large-scale multifaceted dietary benchmark dataset related to opioid users \textit{at the first attempt} and develop a novel framework - i.e., namely \textbf{O}pioid Misuse \textbf{D}etection with \textbf{IN}terpretable \textbf{Diet}ary Patterns (\textbf{Diet-ODIN}) - to bridge heterogeneous graph (HG) and large language model (LLM) for the identification of users with opioid misuse and the interpretation of their associated dietary patterns. Specifically, in Diet-ODIN, based on the National Health and Nutrition Examination Survey (NHANES) {data} \cite{NHANES}, which intends to evaluate the health and nutritional status of adults and children in the United States, we first construct the \textit{NHANES Dietary Graph} with heterogeneous graph structure to comprehensively incorporate users' dietary and health-related information. Then, Diet-ODIN is designed to detect users with potential opioid misuse through their dietary patterns, including food choices and eating habits, and to further provide pertinent interpretations. Nevertheless, distinguishing users with potential opioid misuse from others through their dietary patterns is a non-trivial task, as it involves comprehensively leveraging multifaceted yet complex information, including (1) the common dietary patterns among users, and (2) the specific foods and/or habits adopted by individuals; (3) in addition, noise in real-life dataset, such as users' faulty dietary recollections or data recording errors, renders the task more challenging. To jointly tackle the three grand challenges above for the detection of users with potential opioid misuese, based on the constructed \textit{NHANES Dietary Graph}, Diet-ODIN presents the \textbf{N}oise \textbf{R}educing \textbf{H}eterogeneous \textbf{G}raph \textbf{N}eural \textbf{N}etwork (\textbf{NR-HGNN}) by introducing a holistic graph learning framework to fully integrate both users' individual dietary habits and shared dietary patterns with a refinement component for denoising. Moreover, to interpret the correlation between detected users and associated dietary patterns, Diet-ODIN further bridges LLM and NR-HGNN to provide explicit explanations by utilizing the knowledge learned from the NR-HGNN as prompts for LLM. Last but not least, the new findings based on the extensive analysis performed over the outputs from Diet-ODIN unveil new insights for combating the opioid crisis. Overall, the key contributions of this work are summarized as:

\vspace{-0.1cm}
\begin{itemize}[leftmargin=*]

    \item To the best of our knowledge, this is the \textit{first} AI-driven framework that comprehensively explores the correlations between opioid misuse and users' diets with explainable outputs. In this work, we have also established a large-scale multifaceted dietary benchmark dataset related to opioid users at the first attempt for other researchers and practitioners towards this line of research.

    \item We develop an \textit{innovative and holistic} framework named \textbf{Diet-ODIN} to detect users with potential opioid misuse with interpretable dietary patterns. Based on the builit \textit{NHANES Dietary Graph}, this framework presents a novel \textbf{NR-HGNN} model to jointly capitalize shared dietary patterns among users with a micro-level aggregation module and individual dietary behaviors with a macro-level aggregation module, while also emphasizing noise reduction with a refinement component. Moreover, to interpret the relationship between dietary patterns and opioid misuse, Diet-ODIN implements two innovative strategies leveraging the knowledge learned from \textbf{NR-HGNN} for an LLM to enhance its effectiveness in analytical reasoning.

    \item We conduct extensive experiments based on our established benchmark, whose results show that Diet-ODIN outperforms existing baselines in detecting potential opioid misuse and offers reliable interpretations of the correlations between opioid misuse and dietary patterns, aligning with both existing literature and statistical analysis. These interpretations provide \textit{new insights} for combating the opioid crisis, shedding light on prospective avenues for future research. 

\end{itemize}

\section{Related Work}

\noindent\textbf{Heterogeneous Graph Neural Networks.} Graph Neural Networks (GNNs) have demonstrated effective in a variety of graph learning tasks, outperforming traditional methods~\cite{perozzi2014deepwalk,tang2015line,grover2016node2vec} by leveraging advanced message-passing frameworks ~\cite{kipf2016semi,hamilton2017inductive,velivckovic2017graph}. However, many approaches mainly focus on homogeneous graphs and struggle to interpret the complex semantics inherent in heterogeneous graphs, which consist of diverse node and relation types. To address this, several studies~\cite{dong2017metapath2vec,wang2019heterogeneous,zhang2019heterogeneous,fu2020magnn,wang2021self,wang2023heterogeneous} have utilized meta-paths to capture high-order semantics from such graphs.
Despite their advances, these methods often heavily rely on expert knowledge for meta-path definition and might overlook contextual information. In response, alternative efforts~\cite{schlichtkrull2018modeling,hu2020heterogeneous,lv2021we} have introduced type-specific transformations to better respect the graphs' heterogeneity on simpler, direct connections, but at the risk of underrepresenting longer-range dependencies. Moreover, both approaches may struggle with noise frequently existing in real-world heterogeneous graphs. In contrast, based on the our built our \textit{NHANES Dietary Graph}, NR-HGNN addresses these issues by balancing the integration of both micro- and macro-level information aggregation for richer node embeddings and introducing a refinement module for noise reduction in practical settings.


\noindent\textbf{Large Language Models for Reasoning.} Large language models~\cite{brown2020language,touvron2023llama1,touvron2023llama2} have demonstrated remarkable reasoning capabilities across a range of commonsense tasks~\cite{vaswani2017attention,brown2020language}.
Despite their success, LLMs often encounter challenges in fields that demand specialized knowledge, including healthcare~\cite{yang2023towards}, law~\cite{blair2023can}, finance~\cite{son2023beyond}, and medicine~\cite{tang2023does}. Addressing this limitation, researchers have enriched LLMs with domain-specific knowledge through manual interventions, such as tailored prompts~\cite{tang2023does,son2023beyond} and in-context learning~\cite{yang2023towards,blair2023can}, by elucidating task-specific factors. Drawing on these advancements, we propose to enhance LLMs' reasoning abilities to discern correlations between dietary patterns and opioid usage. Specifically, by utilizing the knowledge learned from the NR-HGNN as prompts, we elegantly bridge LLM and NR-HGNN to interpret the complex interplay between opioid misuse and dietary patterns.

\noindent\textbf{Opioid Misuse Detection.} Research in opioid misuse detection encompasses both quantitative and qualitative approaches. Quantitative studies predominantly employ machine learning algorithms for predicting opioid overdose~\cite{zhang2021rxnet, wen2022disentangled} or relapse risk from medical records~\cite{han2020using, fouladvand2021predicting}, or for identifying opioid users and traffickers through social media data analysis~\cite{qian2021distilling, hu2023fine, fan2018automatic}. 
Nonetheless, these approaches primarily focus on prediction rather than offering detailed analyses or innovative insights into opioid misuse prevention. On the qualitative side, research within healthcare domain examines the link between dietary patterns and chronic opioid usage through case studies and qualitative analyses~\cite{cunningham2016use, whatnall2021efficacy}, highlighting significant correlations~\cite{avena2008evidence, mahboub2021nutritional}. Yet, these studies are often restricted to specific instances and lack a predictive model applicable universally. Addressing this gap, we introduce a comprehensive framework that not only detects but also provides interpretable analyses of opioid misuse based on dietary patterns.

\section{NHANES Dietary Graph}

In this section, we first elucidate the key concepts foundational to our study, followed by an exposition on \textit{NHANES Dietary Graph}.

\begin{definition}[Heterogeneous Graphs] A Heterogeneous Graph (HG)~\cite{wang2022survey} is denoted as $\mathcal{G} = (\mathcal{V}, \mathcal{E}, \mathcal{T}, \mathcal{R})$, comprising sets of nodes $\mathcal{V}$ and edges $\mathcal{E}$. $\mathcal{T}$ and $\mathcal{R}$ represent the types of nodes and edges, respectively, with the condition that $|\mathcal{T}| + |\mathcal{R}| > 2$. Additionally, $\tau(\cdot)$ and $\varphi(\cdot; \cdot)$ serve as mapping functions to identify the types of nodes and edges, where $\tau(i) \in \mathcal{T}$ for any $i \in \mathcal{V}$, and $\varphi(i, j) \in \mathcal{R}$ for any edge $(i,j) \in \mathcal{E}$.
\end{definition}
\begin{definition}[Meta-Path]
    A Meta-path $m \in \mathcal{M}$~\cite{wang2019heterogeneous} represents a sequence of node and edge types, functioning as a composite of relations: $\mathcal{P}_m: \mathcal{T}_1 \xrightarrow{\mathcal{R}_1} \mathcal{T}_2 \xrightarrow{\mathcal{R}_2} \ldots \xrightarrow{\mathcal{R}_{n-1}} \mathcal{T}_n$. It encapsulates a particular form of high-order semantics within an HG, with different meta-paths potentially offering complementary insights.
\end{definition}
\begin{definition}[Meta-Path-based Neighborhoods]
    Meta-path-based neighborhoods refer to sets of neighborhoods interconnected through specific meta-paths. We define $\mathcal{P}_m(i,j)$ as the function measuring the path count connecting nodes $i$ and $j$ via meta-path $m$, thereby delineating these neighborhoods as $\mathcal{N}_{\mathcal{P}_m}^k(i) = \{j | \mathcal{P}_m(i,j) \ge k\}$, typically with $k = 1$.
\end{definition}

\begin{figure}[!t]
  \centering
  \includegraphics[width=\linewidth]{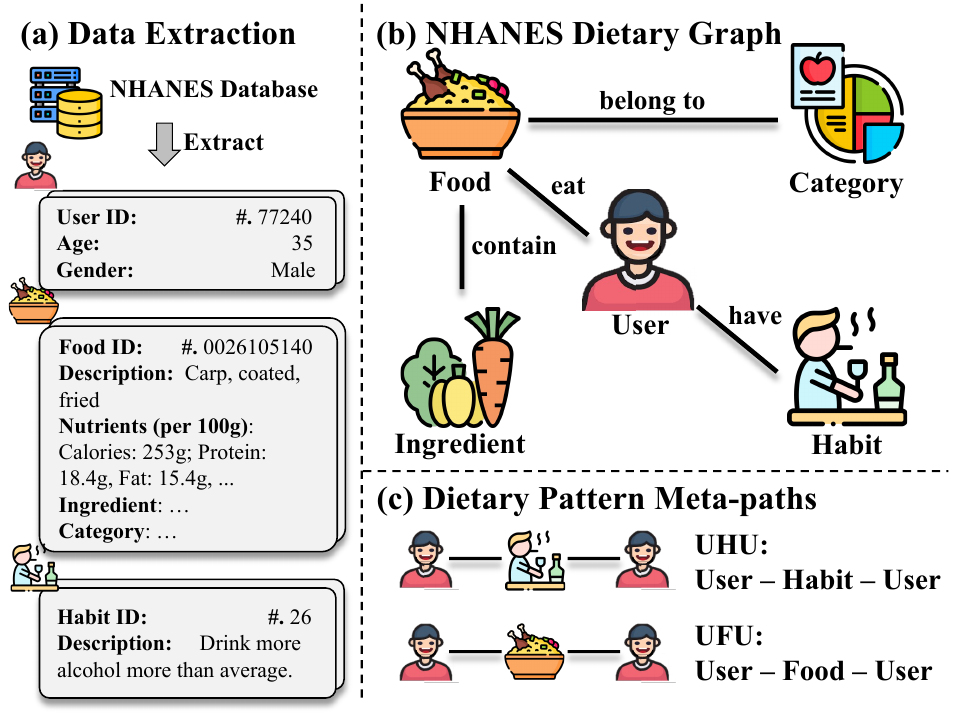}
  \vspace{-20pt}
  \caption{The schema of NHANES Dietary Graph.}
  \vspace{-10pt}
  \label{fig:dietary graph}
\end{figure}

\begin{figure*}[!t]
  \centering
  \vspace{-5pt}
  \includegraphics[width=1.0\textwidth]{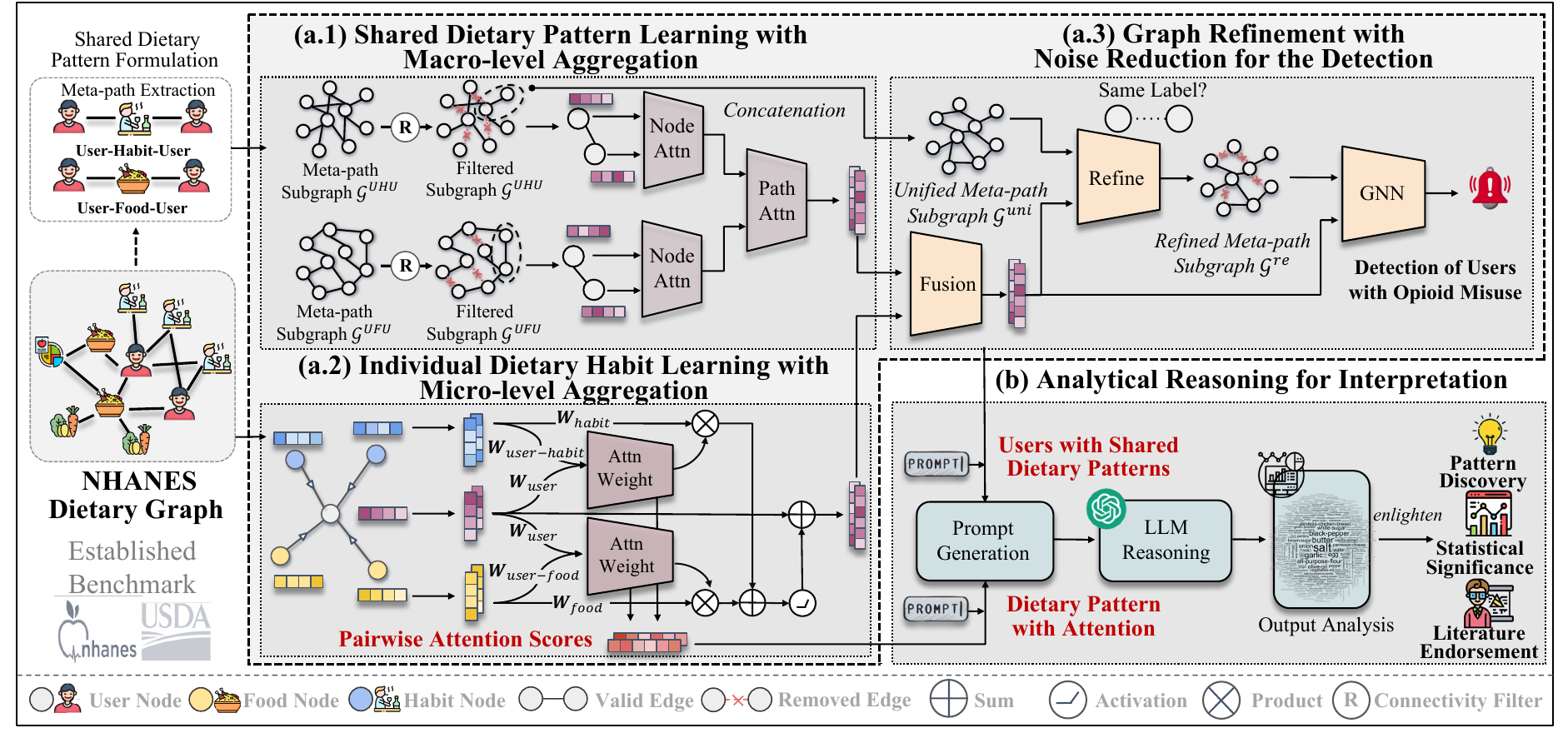}
  \vspace{-20pt}
  \caption{The overview framework of Diet-ODIN, which consists of (a) a graph learning framework called NR-HGNN for detecting opioid users, and (b) an LLM-powered reasoning module for interpreting the most important dietary patterns. 
  }
  \vspace{-10pt}
  \label{fig:main_framework}
\end{figure*}

\noindent\textbf{Constructing NHANES Dietary Graph.} Utilizing NHANES data from 2003 to 2020, we developed NHANES Dietary Graph, the first graph dataset aiming at investigating the relationship between dietary patterns and opioid misuse. This graph is constructed from data on food intake, dietary habits, and prescription drug usage. As depicted in Figure \ref{fig:dietary graph}, it features five types of nodes: \textit{user}, \textit{food}, \textit{habit}, \textit{ingredient}, and \textit{category}, and four types of relationships: \textit{user-eat-food}, \textit{user-have-habit}, \textit{food-contain-ingredient}, and \textit{food-belong\_to-category}. Within this graph, we can difine different meta-paths, such as \textit{User-Habit-User (UHU)} and \textit{User-Food-User (UFU)}, which represent shared dietary habits and food consumption patterns among users, respectively. Opioid users are identified by either (1) records of illicit opioid drug use, like heroin, within a year, or (2) records of prescription opioid medication use for over 90 days, which is a criteria commonly employed in
the domain ~\cite{gu2022prevalence}. Thus, we form the opioid user detection as a binary classification task. User features include anonymized demographic information, while food features comprise nutritional values, such as calories, protein, and sodium. For text-described node types such as food, habit, ingredient, and category, we employ pre-trained BERT~\cite{devlin2018bert} to encode their descriptions into node features. Details on feature processing and dataset statistics are available in Appendix \ref{appendix:Graph Construction Details}.

\section{The Framework of Diet-ODIN}

Based on the built \textit{NHANES Dietary Graph}, we introduce an innovative framework for \textbf{O}pioid Misuse \textbf{D}etection with \textbf{IN}terpretable \textbf{Diet}ary Patterns (\textbf{Diet-ODIN}), illustrated in Figure \ref{fig:main_framework}. The framework incorporates a \textbf{N}oise \textbf{R}educing \textbf{H}eterogeneous \textbf{G}raph \textbf{N}eural \textbf{N}etwork (\textbf{NR-HGNN}) aimed at identifying users with opioid misuse (detailed in Section \ref{sec:method detection}), and a LLM-based interpretation module  bridged with NR-HGNN to explore the relationship between dietary patterns and opioid misuse (outlined in Section \ref{sec:method reasoning}).

\subsection{NR-HGNN for Detecting Users with Opioid Misuse}
\label{sec:method detection}

Detecting users with opioid misuse presents a complex challenge, requiring comprehensive analysis of (1) shared dietary patterns among users, (2) individual food consumption or habits, and (3) reducing noise introduced in data collection. Our NR-HGNN model addresses these issues by jointly aggregating the information at both macro-level (to tackle the first challenge) and micro-level (to address the second challenge) and performing noise reduction through graph refinement strategies (to resolve the third challenge). As illustrated in Figure \ref{fig:main_framework} (a), the NR-HGNN model comprises three principal components: (1) shared dietary pattern learning with marco-level aggregation (i.e., Figure \ref{fig:main_framework} (a.1)), (2) individual dietary habit learning with micro-level aggregation  (i.e., Figure \ref{fig:main_framework} (a.2)), and (3) graph refinement with noise reduction for the detection  (i.e., Figure \ref{fig:main_framework} (a.3)), with detailed discussions to follow.



\subsubsection{Shared Dietary Pattern Learning with Marco-level Aggregation}

Users with opioid misuse often exhibit unique dietary preferences, notably towards sugar and high-calorie foods, distinguishing them from non-opioid users~\cite{Morabia_Fabre_Ghee_Zeger_Orsat_Robert_1989,Mysels_Sullivan_2010}. Motivated by this observation, we introduce macro-level aggregation to utilize shared dietary patterns between users for the detection of opioid misuse. Specifically, we employ meta-paths \textit{User-Habit-User (UHU)} and \textit{User-Food-User (UFU)} to capture the shared food consumption and habit patterns among users. Following the methodology of \cite{wang2019heterogeneous}, these meta-paths facilitate the extraction of subgraphs by linking users with identical dietary patterns, thereby preserving intricate high-order semantics. Nevertheless, direct extraction may lead to the inclusion of noisy subgraphs, as dietary patterns are not solely determined by individual foods or habits~\cite{avena2008evidence, mahboub2021nutritional}. For example, deducing similar dietary habits from a single shared meal, such as fried chicken, is over-simplistic. 
Addressing this, we introduce a novel subgraph extraction method, termed \textit{connectivity filtering}, which establishes connections between two users only if they share multiple dietary habits or foods:
\begin{equation}
\mathcal{G}^m = (\mathcal{V}^m, \mathcal{E}^m), \mathcal{E}^m = \{ (i,j) | j \in \mathcal{N}_{\mathcal{P}_m}^k(i) \},
\end{equation}
where $\mathcal{G}^m$ represents the subgraph generated through meta-path $\mathcal{P}_m$, and $k$ signifies an empirically determined threshold for selection. This methodology ensures the retention of only the most pertinent connections within the subgraphs.

Following the establishment of meta-path-induced subgraphs, we delve into the extraction of semantics from each meta-path, effectively modeled through message-passing mechanisms:
\begin{equation}
    \textstyle \mathbf{h}_i^m = \sum_{j \in \mathcal{N}^k_{\mathcal{P}m}(i)} \alpha_{ij}^m \cdot \mathbf{W} \mathbf{h}_j^m,
\end{equation}
where $\mathbf{h}^m$ represents node embeddings within subgraph $\mathcal{G}^m$, $\mathbf{W}$ denotes a learnable matrix, and $\alpha_{ij}^m$ signifies the correlation between nodes $i$ and $j$, which is learnable or fixed. To capture fine-grained knowledge, we further attention mechanisms to dynamically allocate these correlations:
\begin{equation}
    \alpha_{ij}^m = \frac{\exp ( \sigma ( \mathbf{a}^\top \cdot [\mathbf{W} \mathbf{h}_i^m \;\|\; \mathbf{W} \mathbf{h}_j^m] ) )}{\sum_{t \in \mathcal{N}^k_{\mathcal{P}_m}(i)} \exp(\sigma(\mathbf{a}^\top \cdot [\mathbf{W} \mathbf{h}_i^m \;\|\; \mathbf{W} \mathbf{h}_t^m]))},
\end{equation}
with $\sigma(\cdot)$ as the activation function and $\mathbf{a}$ as the learnable attention vector. Recognizing the complex etiology behind opioid misuse, the attention also evaluates the significance of various factors (e.g., food and habits) in prediction. Integration of insights from different meta-paths (\textit{UFU} and \textit{UHU}) involves analyzing their importance:
\begin{equation}
    \textstyle \mathbf{h}_i^{Ma} = \sigma(\sum_{m \in \mathcal{M}} \beta^{m} \cdot \mathbf{h}_i^m),
\end{equation}
\begin{equation}
    \beta^m = \frac{\exp (\sum_{i \in \mathcal{V}^m} \mathbf{q}^\top \cdot \tanh (\mathbf{W} \mathbf{h}_i)) }{\sum_{m \in \mathcal{M}} \exp (\sum_{i \in \mathcal{V}^m} \mathbf{q}^\top \cdot \tanh (\mathbf{W} \mathbf{h}_i))},
\end{equation}
where $\mathbf{h}^{Ma}$ encapsulates the macro-level node embeddings, and $\mathbf{q}$ is an additional attention vector that discerns the relevance of various meta-paths. This hierarchical attention mechanism empowers the model to discern nuanced high-order semantics within the graph.

\subsubsection{Individual Dietary Habit Learning with Micro-level Aggregation}
\label{sec:con aggr}

Macro-level aggregation captures high-order information within HGs, yet may overlook crucial localized knowledge essential for accurate opioid misuse identification~\cite{cunningham2016use, whatnall2021efficacy}. This localized information, embodying the direct relationships between users and their dietary behaviors—specifically, their food intake and habits—varies in its prediction for opioid misuse. To incorporate this low-order information, we introduce a micro-level aggregation module. This module is designed to discern the relevance of particular dietary patterns dynamically, thereby facilitating prediction accuracy:
\begin{equation}
    \textstyle \mathbf{h}_i^{Mi} = \sigma(\frac{1}{| \mathcal{N}(i) |} \sum_{j \in \mathcal{N}(i)} \alpha_{ij}^{Mi} \cdot \mathbf{W}_{\tau(j)} \mathbf{h}_j^{Mi} ),
\end{equation}
where $\mathbf{h}^{Mi}$ represents the micro-level node embedding, $\mathcal{N}(i)$ indicates node $i$'s direct neighbors, and $\mathbf{W}_{\tau(j)}$ applies node-type specific transformations reflecting HG's inherent heterogeneity. The micro-level attention $\alpha_{ij}^{Mi}$ quantifies each neighbor's significance, capturing the dietary pattern-user relationship:
\begin{equation}
    \alpha_{ij}^{Mi} = \softmax \left([\mathbf{W}_{\tau(i)} \mathbf{h}_i^{Mi}]^\top \cdot \mathbf{W}_{\varphi(i,j)} \cdot [\mathbf{W}_{\tau(j)} \mathbf{h}_j^{Mi}] / \sqrt{d} \right),
\end{equation}
$\mathbf{W}_{\tau(\cdot)}$ and $\mathbf{W}_{\varphi(\cdot, \cdot)}$ enable precise modeling of node and relation types, enhancing the framework's ability to navigate HG's complexity. The introduction of a normalization factor $\sqrt{d}$ in attention scoring mitigates potential gradient vanishing and stabilizes training~\cite{vaswani2017attention}. Beyond enhancing prediction capabilities, this attention-based method prioritizes specific dietary behaviors in the reasoning process, as further discussed in Section \ref{sec:method reasoning}.

\subsubsection{Graph Refinement with Noise Reduction for the Detection}

The macro-level aggregation captures high-order information while micro-level aggregation processes low-order information, both of which are essential and complementary aspects of our NR-HGNN. To integrate these embeddings and fully leverage the information within the HG, we apply the following fusion operation:
\begin{equation}
\mathbf{h}_i = \text{ReLU} (\mathbf{W} \cdot [ \mathbf{h}_i^{Ma} \;\|\; \mathbf{h}_i^{Mi} ] + \mathbf{b}).
\end{equation}
However, empirical evidence suggests that directly applying the fused embedding $\mathbf{h}$ for prediction leads to sub-optimal. We attribute this to potential noise from data collection processes, where inaccuracies may arise from users' faulty recollections of dietary intake or frequency of certain habits, as well as possible errors introduced by interviewers. Such inaccuracies can manifest as either spurious or missing connections in the graph, consequently generating noise in the graph structure. This noise can render the embeddings derived from the two aggregation modules unreliable by falsely correlating graph structure with predictions. To address this, we propose initially refining the noisy dietary patterns within the graph structure. Subsequently, we input both the node embeddings and the refined graph into an additional GNN to derive more accurate embeddings.

We consider leveraging user correlations for refinement, positing that users with similar dietary patterns are likely to exhibit similarities. To consolidate various yet complementary dietary patterns, we integrate different meta-paths into a unified graph:
\begin{equation}
    \textstyle \mathcal{G}^{uni} = \bigcup_{m \in \mathcal{M}} \mathcal{G}^m = (\mathcal{V}^{uni}, \mathcal{E}^{uni}),
\end{equation}
where $\mathcal{M}$ represents the set of meta-paths, including \textit{User-Food-User} and \textit{User-Habit-User}. To determine the refinement signal, we consider if two connected users share the same label, the connection would be reliable, based on the hypothesis that dietary patterns of opioid and non-opioid users differ significantly. This leads us to model refinement through a link prediction approach:
\begin{equation}
    \mathcal{L}_{refine} = - \frac{1}{|\mathcal{E}^{uni}_{train}|} \sum_{(i,j) \in \mathcal{E}^{uni}_{train}} y_{ij} \log (\hat{y_{ij}}) + (1 - y_{ij}) \log (1 - \hat{y_{ij}}),
\end{equation}
where $y_{ij} = pred_{refine}([\mathbf{h}_i  \|  \mathbf{h}_j])$ predicts the edge reliability. A subset of edges, $\mathcal{E}^{uni}_{train}$, is selected for training the refinement model, ensuring it only accesses nodes from the training dataset to prevent information leakage from unseen nodes. Subsequently, the refinement model evaluates all edges in $\mathcal{G}^{uni}$ for reliability, discarding those deemed unreliable:
\begin{equation}
    \label{eq:refined graph}
    \mathcal{G}^{re} = (\mathcal{V}^{uni}, \mathcal{E}^{re}), \mathcal{E}^{re} = \{ (i,j) \in \mathcal{E}^{uni} | \hat{y_{ij}} = 1 \}.
\end{equation}
This process enhances graph homophily, a crucial aspect in analyzing social phenomena between people~\cite{ertug2022does}. The refined graph is then employed to derive reliable node embeddings for predictions:
\begin{equation}
    \hat{\mathbf{h}_i} = \phi(\mathbf{h}_i, \;\mathcal{G}^{re}),
\end{equation}
\begin{equation}
    \hat{y_i} = pred_{node}(\hat{\mathbf{h}_i}) = \softmax(\mathbf{W} \cdot \hat{\mathbf{h}i} + \mathbf{b}),
\end{equation}
where $\phi(\cdot, \cdot)$ denotes the GNN for refining node embeddings $\hat{\mathbf{h}}$, and $\hat{y_i}$ represents the model predictions. Optimization is achieved through minimizing the cross-entropy loss:
\begin{equation}
\mathcal{L} = - \frac{1}{ |\mathcal{V}| } \sum_{i \in \mathcal{V}} y_i \log(\hat{y_i}) + (1 - y_i) \log (1 - \hat{y_i}).
\end{equation}
As refined graphs maintain more reliable patterns, they also enhance the reasoning process, as elucidated in Section \ref{sec:method reasoning}.

\begin{figure}[!t]
  \centering
  \includegraphics[width=\linewidth]{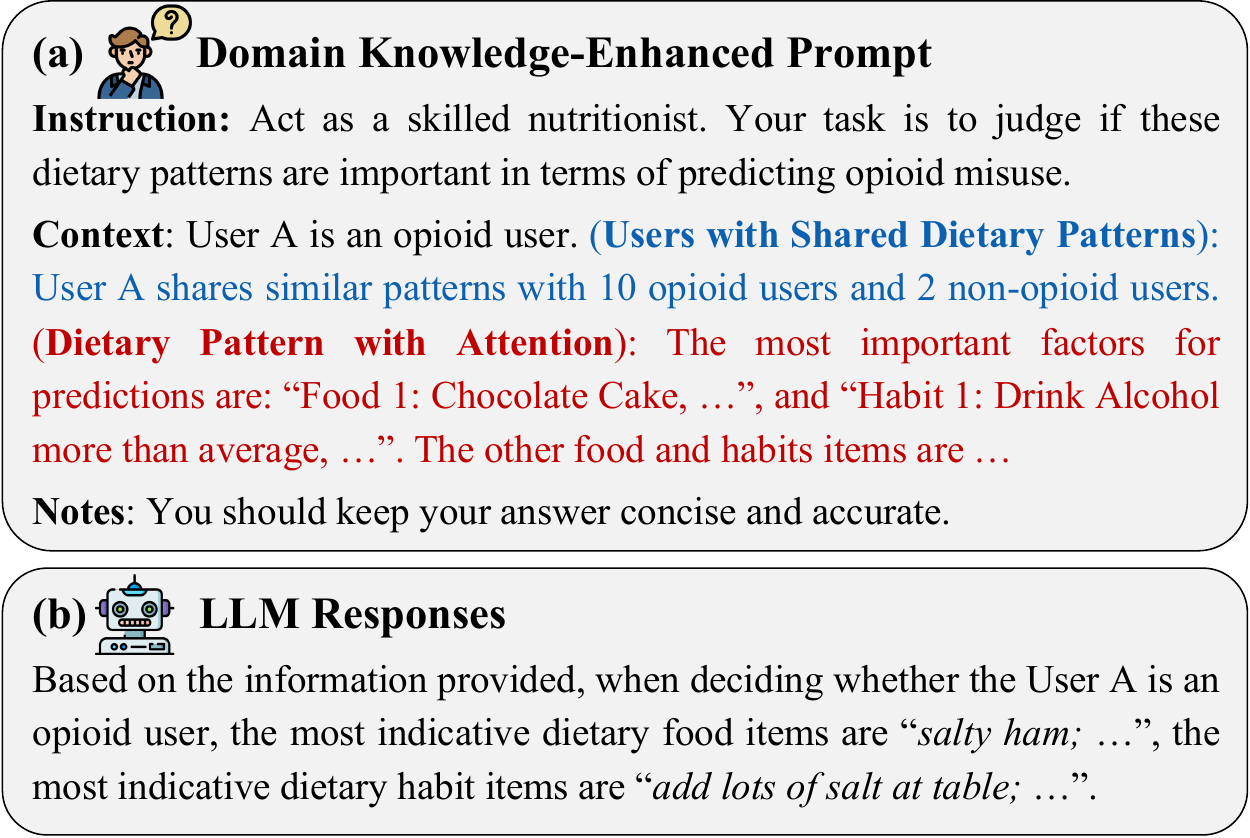}
  \vspace{-20pt}
  \caption{The prompts (highlighted in blue and red) generated from the NR-HGNN for interpreting key dietary patterns that indicate opioid misuse in individuals.}
  \vspace{-20pt}
  \label{fig:prompt}
\end{figure}

\subsection{Bridging NR-HGNN and LLM for Interpretation}
\label{sec:method reasoning}

Beyond detecting users with opioid misuse, our Diet-ODIN framework also excels at interpreting key dietary patterns associated with opioid misuse, thereby aiding in the prevention and treatment of its misuse. To achieve this, Diet-ODIN bridges NR-HGNN and LLM to investigate the nexus between opioid misuse and dietary behaviors, providing individuals with meaningful insights. To this end, our objective is to design prompts that enable LLM to perform trustworthy reasoning. As demonstrated in Figure \ref{fig:prompt}, a prompt consists of three parts: (1) \textit{Instruction}, (2) \textit{Context}, and (3) \textit{Notes}. The \textit{Instruction} conveys task-specific directives to the LLM, clarifying the task at hand. The \textit{Context} provides user-specific information, facilitating tailored reasoning. The \textit{Notes} are employed to guide LLM behavior, such as restricting output length. Notably, \textit{Context} is the most crucial component, presenting user information. A straightforward context design is to directly convert the characteristics of users and their dietary patterns in text, satisfying the input format of LLM; but the method cannot fully exploit the potential of LLM, particularly in domains like healthcare, where domain-specific knowledge is vital~\cite{tang2023does}. To address this, recent studies~\cite{yang2023towards,blair2023can} have incorporated domain-specific knowledge into prompt generation, thereby enriching the information available for reasoning. Inspired by these works, we introduce two strategies to boost the reasoning power of LLM for our task.

\begin{table*}[!t]
    \vspace{0pt}
    \caption{Performance comparison between NR-HGNN and existing methods for identifying users with opioid misuse. Results are reported as mean ± std\%. The best performance is bolded and runner-ups are underlined.}
    \vspace{-10pt}
    \resizebox{\linewidth}{!}{
        \begin{tabular}{l  | ccccc | ccccc}
            \toprule
                                                    & \multicolumn{5}{c|}{\bf Train 20\% $-$ Valid 40\% $-$ Test 40\%} & \multicolumn{5}{c}{\bf Train 40\% $-$ Valid 30\% $-$ Test 30\%}                                                                                                                                                                                                                                                                         \\
            Methods                                 & Accuracy                                                     & Precision                                                   & Recall                         & F1-Score                       & ROC-AUC                        & Accuracy                       & Precision                      & Recall                         & F1-Score                       & ROC-AUC                        \\ \midrule
            MLP                                     & 70.15\small{±0.18}                                           & 68.31\small{±0.69}                                          & 75.19\small{±0.17}             & 71.57\small{±0.45}             & 77.52\small{±0.08}             & 70.40\small{±0.33}             & 68.33\small{±0.94}             & 76.13\small{±0.20}             & 72.00\small{±0.40}             & 77.38\small{±0.04}             \\
            GCN\cite{kipf2016semi}                  & 73.26\small{±0.24}                                           & 69.87\small{±0.20}                                          & 81.72\small{±0.61}             & 75.33\small{±0.28}             & 78.09\small{±0.12}             & 72.94\small{±0.21}             & 68.98\small{±0.18}             & 83.36\small{±0.54}             & 75.49\small{±0.24}             & 78.10\small{±0.12}             \\
            SAGE\cite{hamilton2017inductive}        & 73.93\small{±0.25}                                           & 70.08\small{±0.27}                                          & 83.46\small{±0.56}             & 76.19\small{±0.25}             & 78.62\small{±0.14}             & 73.11\small{±0.27}             & 69.05\small{±0.17}             & 83.77\small{±0.56}             & 75.70\small{±0.30}             & 78.50\small{±0.11}             \\
            GAT\cite{velivckovic2017graph}          & 75.06\small{±0.12}                                           & 69.96\small{±0.12}                                          & 87.78\small{±0.36}             & 77.87\small{±0.14}             & 80.93\small{±0.02}             & 74.54\small{±0.06}             & 69.57\small{±0.05}             & 87.24\small{±0.17}             & 77.41\small{±0.07}             & 80.33\small{±0.02}             \\ \midrule
            RGCN\cite{schlichtkrull2018modeling}    & 75.64\small{±0.09}                                           & 70.15\small{±0.21}                                          & 89.24\small{±0.37}             & 78.55\small{±0.03}             & \underline{81.27\small{±0.02}} & \underline{74.82\small{±0.17}} & 69.82\small{±0.10}             & 87.43\small{±0.34}             & 77.64\small{±0.18}             & 80.93\small{±0.03}             \\
            HGT\cite{hu2020heterogeneous}           & \underline{75.72\small{±0.06}}                               & 69.38\small{±0.06}                                          & \underline{92.02\small{±0.01}} & \underline{79.11\small{±0.04}} & 81.14\small{±0.01}             & 74.62\small{±0.05}             & 68.96\small{±0.04}             & \underline{89.54\small{±0.07}} & \underline{77.92\small{±0.04}} & 80.69\small{±0.01}             \\
            HAN\cite{wang2019heterogeneous}         & 75.10\small{±0.02}                                           & 70.37\small{±0.01}                                          & 86.64\small{±0.03}             & 77.66\small{±0.02}             & 80.02\small{±0.01}             & 74.26\small{±0.03}             & 69.66\small{±0.02}             & 85.95\small{±0.07}             & 76.96\small{±0.04}             & 80.52\small{±0.01}             \\
            SHGN\cite{lv2021we}                     & 73.92\small{±1.20}                                           & \underline{70.51\small{±1.33}}                              & 82.34\small{±4.23}             & 75.90\small{±1.59}             & 80.80\small{±0.85}             & 73.74\small{±1.13}             & \underline{71.20\small{±0.89}} & 79.77\small{±2.97}             & 75.21\small{±1.43}             & \underline{81.06\small{±0.68}} \\
            SeHGNN\cite{yang2023simple}             & 75.48\small{±0.17}                                           & 70.00\small{±0.20}                                          & 89.16\small{±0.60}             & 78.42\small{±0.20}             & 81.23\small{±0.18}             & 74.63\small{±0.22}             & 69.63\small{±0.23}             & 87.35\small{±0.59}             & 77.49\small{±0.23}             & 80.84\small{±0.19}             \\
            \rowcolor{Gray} \textbf{NR-HGNN (Ours)} & \textbf{91.23\small{±0.63}}                                  & \textbf{88.55\small{±1.53}}                                 & \textbf{94.74\small{±1.19}}    & \textbf{91.52\small{±0.54}}    & \textbf{96.50\small{±0.44}}    & \textbf{91.68\small{±0.24}}    & \textbf{89.65\small{±0.52}}    & \textbf{94.24\small{±0.60}}    & \textbf{91.89\small{±0.23}}    & \textbf{96.62\small{±0.34}}    \\ \bottomrule
        \end{tabular}
    }
    \vspace{-5pt}
    \label{tab:results}
\end{table*}

\begin{table*}[!t]
    \caption{Performance comparison with different variants of our proposed NR-HGNN, including (a.1) macro-level aggregation, (a.2) micro-level aggregation, and (a.3) graph refinement with noise reduction.}
    \vspace{-10pt}
    \resizebox{\linewidth}{!}{
        \begin{tabular}{ccc | ccccc | ccccc}
            \toprule
                                    &               &              & \multicolumn{5}{c|}{\bf Train 20\% $-$ Valid 40\% $-$ Test 40\%} & \multicolumn{5}{c}{\bf Train 40\% $-$ Valid 30\% $-$ Test 30\%}                                                                                                                                                                                                                                                                         \\
            \textbf{(a.1)}           & \textbf{(a.2)} & \textbf{(a.3)} & Accuracy                                                     & Precision                                                   & Recall                         & F1-Score                       & ROC-AUC                        & Accuracy                       & Precision                      & Recall                         & F1-Score                       & ROC-AUC                        \\ \midrule
            $\surd$                 & $-$           & $-$          & 84.75\small{±0.72}                                           & 82.60\small{±0.56}                                          & 88.03\small{±1.11}             & 85.22\small{±0.74}             & 91.41\small{±0.42}             & 85.08\small{±0.62}             & 82.90\small{±0.58}             & 88.38\small{±1.04}             & 85.55\small{±0.64}             & 91.53\small{±0.35}             \\
            $-$                     & $\surd$       & $-$          & 73.96\small{±0.94}                                           & 72.23\small{±1.59}                                          & 78.12\small{±6.22}             & 74.90\small{±2.07}             & 81.39\small{±0.72}             & 75.30\small{±0.86}             & 73.01\small{±1.04}             & 80.43\small{±4.64}             & 76.45\small{±1.65}             & 82.35\small{±0.16}             \\
            $\surd$                 & $\surd$       & $-$          & \underline{89.46\small{±0.43}}                               & \underline{87.50\small{±1.51}}                              & 92.12\small{±1.89}             & \underline{89.72\small{±0.42}} & \underline{95.01\small{±0.14}} & \underline{88.84\small{±0.77}} & \underline{86.74\small{±2.20}} & 91.84\small{±2.54}             & \underline{89.16\small{±0.67}} & \underline{95.47\small{±0.13}} \\
            $\surd$                 & $-$           & $\surd$      & 86.99\small{±0.70}                                           & 80.79\small{±0.95}                                          & \textbf{97.06\small{±0.45}}    & 88.18\small{±0.57}             & 92.69\small{±0.45}             & 88.34\small{±0.82}             & 83.65\small{±1.31}             & \textbf{95.36\small{±0.71}}    & 89.11\small{±0.68}             & 93.35\small{±0.50}             \\
            $-$                     & $\surd$       & $\surd$      & 70.91\small{±10.37}                                          & 65.65\small{±9.45}                                          & 94.74\small{±3.05}             & 77.04\small{±5.73}             & 76.28\small{±12.1}             & 60.02\small{±8.71}             & 56.92\small{±8.20}             & \underline{95.14\small{±3.42}} & 70.77\small{±4.58}             & 65.01\small{±9.51}             \\
            \rowcolor{Gray} $\surd$ & $\surd$       & $\surd$      & \textbf{91.23\small{±0.63}}                                  & \textbf{88.55\small{±1.53}}                                 & \underline{94.74\small{±1.19}} & \textbf{91.52\small{±0.54}}    & \textbf{96.50\small{±0.44}}    & \textbf{91.68\small{±0.24}}    & \textbf{89.65\small{±0.52}}    & 94.24\small{±0.60}             & \textbf{91.89\small{±0.23}}    & \textbf{96.62\small{±0.34}}    \\ \bottomrule
        \end{tabular}
    }
    \vspace{-5pt}
    \label{tab:ablation}
\end{table*}

\textbf{\textit{Strategies for facilitating domain-specific reasoning.}}
We identify two primary limitations preventing advanced reasoning capability. Firstly, LLMs may not distinguish among various types of dietary patterns (foods and habits) when examining their correlation with opioid misuse. This issue could derive from the lack of domain knowledge of the relationship between specific dietary patterns and opioid misuse, rendering LLMs unable to identify specific foods or habits crucial for detecting opioid misuse. 
Secondly, the presence of outliers among users poses another challenge. For instance, certain opioid users might exhibit dietary patterns more commonly associated with non-opioid users, misleading LLMs in interpreting key dietary patterns of opioid misuse. 

To address the first limitation, we introduce \textit{Dietary Pattern with Attention}, which aims to help the LLM leverage critical dietary patterns that NR-HGNN learns in detecting users with opioid misuse. Specifically, we utilize attention scores retrieved in Micro-level Aggregation (Section \ref{sec:con aggr}) to rank foods and habits according to their importance. We empirically select the top-$10$ foods and habits for each user. By highlighting the importance of specific dietary patterns, we reduce the uncertainty of the LLM, thereby enhancing the reliability of their final interpretations. For the second limitation, we introduce the \textit{Users with Shared Dietary Patterns} strategy to assess the typicality of users based on shared dietary patterns. This strategy aids in determining if dietary patterns accurately indicate opioid usage. In particular, we define typical users as individuals who share patterns with a substantial number of peers holding the same labels. We evaluate the user typicality on the refined graph (Equation \ref{eq:refined graph}), as it preserves more reliable patterns. 
In this way, we avoid the LLM to capture the pseudo-correlation between dietary patterns and interpretations. These two strategies complement each other, which enables the LLM offering more accurate analyses by considering both individual and shared dietary patterns.

\section{Experiments}

\subsection{Experiment Setup}

\noindent\textbf{Baselines.} To demonstrate the superiority of NR-HGNN in detecting users with opioid misuse, we select a range of graph learning baselines. These include vanilla MLP, homogeneous GNNs such as GCN~\cite{kipf2016semi}, GraphSAGE (SAGE)~\cite{hamilton2017inductive}, and GAT~\cite{velivckovic2017graph}, alongside heterogeneous GNNs like RGCN~\cite{schlichtkrull2018modeling}, HAN~\cite{wang2019heterogeneous}, HGT~\cite{hu2020heterogeneous}, and the state-of-the-art Simple-HGN (SHGN)~\cite{lv2021we} and SeHGNN~\cite{yang2023simple}. 

\noindent\textbf{Evaluation Settings.} In this study, we evaluate two settings with distinct training, validation, and testing splits. Specifically, 20\% and 40\% of nodes are selected randomly to create the training sets, with the remainder being evenly divided into validation and testing sets. Each model is executed on ten consecutive seeds to reduce the influence of randomness. To thoroughly assess the model's performance across various dimensions, we employ five widely recognized metrics: Accuracy, Precision, Recall, F1-Score, and ROC-AUC score. For a balanced comparison, all methods are configured with a hidden dimension of 256 and utilize the Adam optimizer with a learning rate of 0.001 and a weight decay of 0.001. We fix the total number of epochs at 500. For baseline methods that incorporate an attention mechanism, the number of attention heads is set to four.

\subsection{Results}

\noindent\textbf{Performance Comparison.} Table \ref{tab:results} demonstrates the performance of baselines and our NR-HGNN in detecting users with opioid misuse on the proposed \textit{NHANES Dietary Graph}, where our NR-HGNN notably surpasses existing methods, including the state-of-the-art SeHGNN.
This substantial performance improvement is attributed to the model's adeptness at concurrently capturing low and high-level information with micro- and macro-level aggregations from the graph, along with graph refinement for noise reduction. Contrary to expectations, increasing training data results in a slight decrease in the performance of most baselines, which might stem from potential over-fitting, where additional data may introduce redundant information. Nevertheless, our model performs better on the 40\% training set split, which indicates the robustness of our methods. Notably, conventional methods demonstrate limitations in accurately identifying non-opioid users, as indicated by high recall but low precision values. This discrepancy may arise from these models accentuating patterns common to both opioid and non-opioid users. In contrast, our NR-HGNN addresses this challenge through its graph refinement process, thereby enhancing the reliability of patterns used for detecting users with opioid misuse.



\noindent\textbf{Ablation Study.} Our method performs significantly better than other baselines. Therefore, to discern the elements contributing to such performance enhancement, we have conducted a comprehensive ablation study, as depicted in Table \ref{tab:ablation}. Our observations reveal that each component independently enhances the model's proficiency in detecting users with opioid misuse. When these three components are integrated, the model demonstrates optimal performance, highlighting the synergistic effect of concurrently processing high-order and low-order information alongside noise reduction refinement. Notably, employing either low-order or high-order information in isolation results in a diminished effectiveness of the refinement process. This outcome underscores the importance of synergistically aggregating information at both micro- and macro- levels to capture diverse facets of the graph, thereby ensuring the reliability of the refined patterns.



\begin{figure}[!ht]
    \centering
    \vspace{-10pt}
    \subfloat[GCN]{\includegraphics[width=0.5\linewidth]{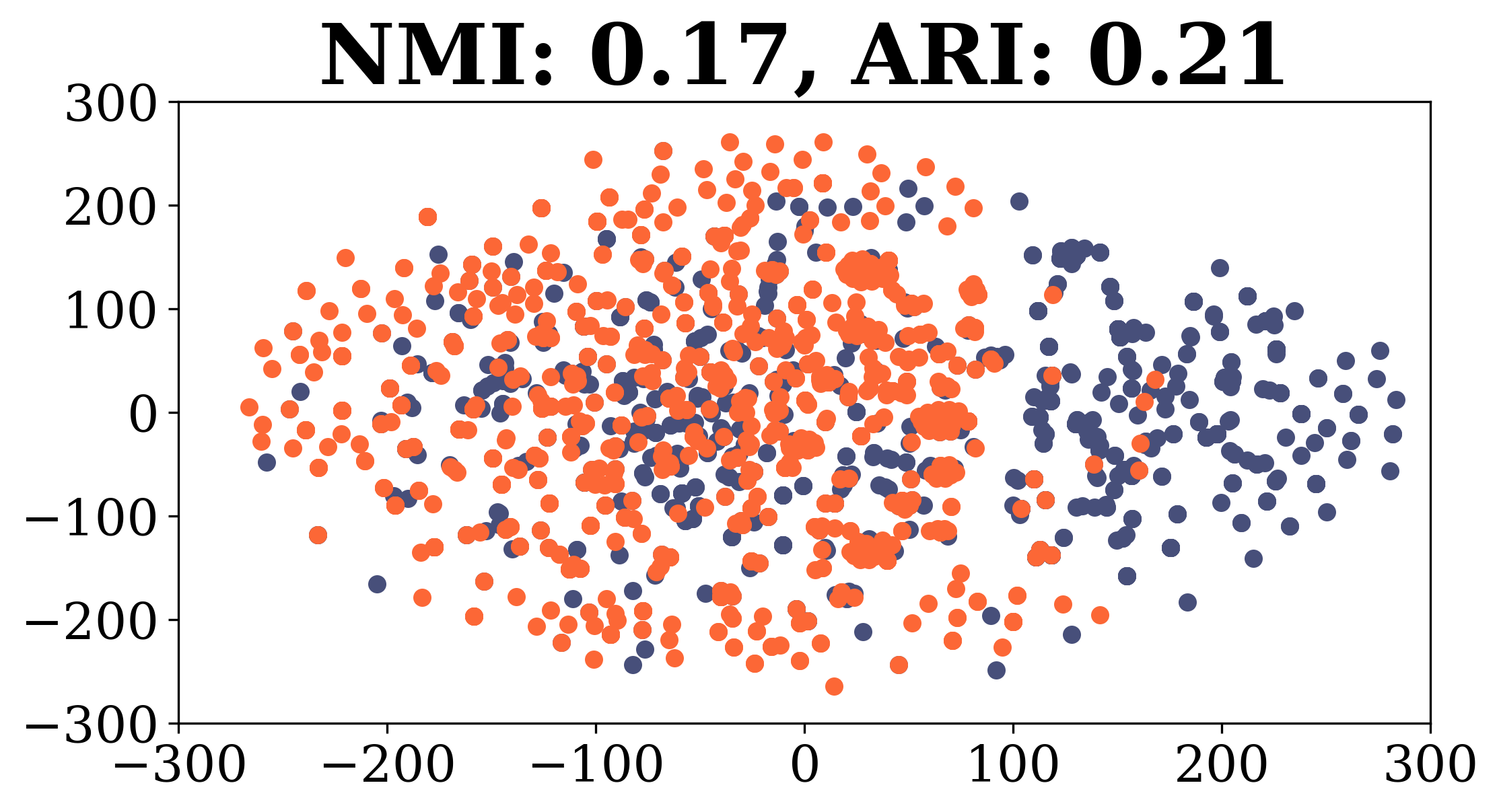}}
    \subfloat[RGCN]{\includegraphics[width=0.5\linewidth]{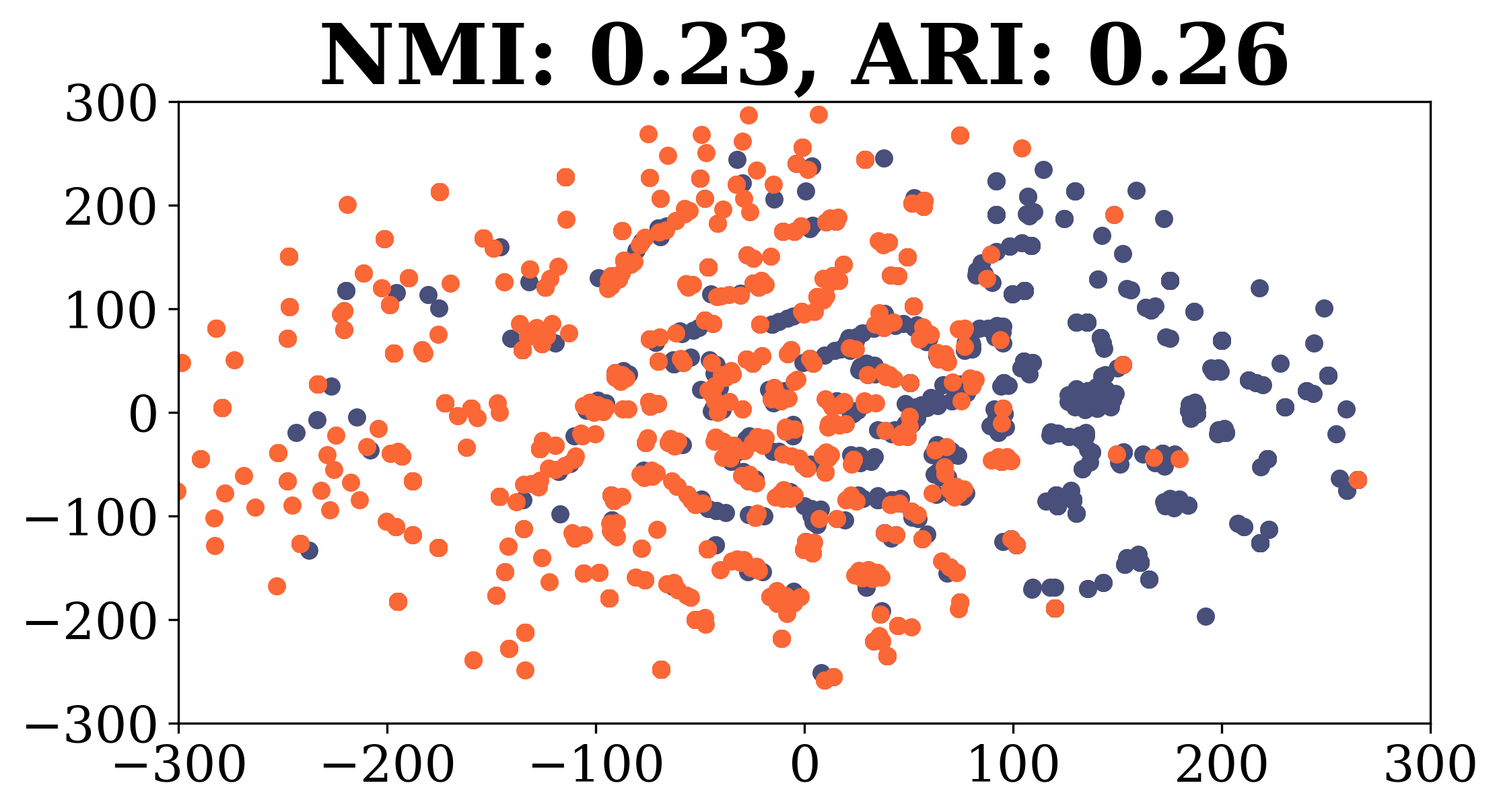}} \\
    \vspace{-10pt}
    \subfloat[SeHGNN]{\includegraphics[width=0.5\linewidth]{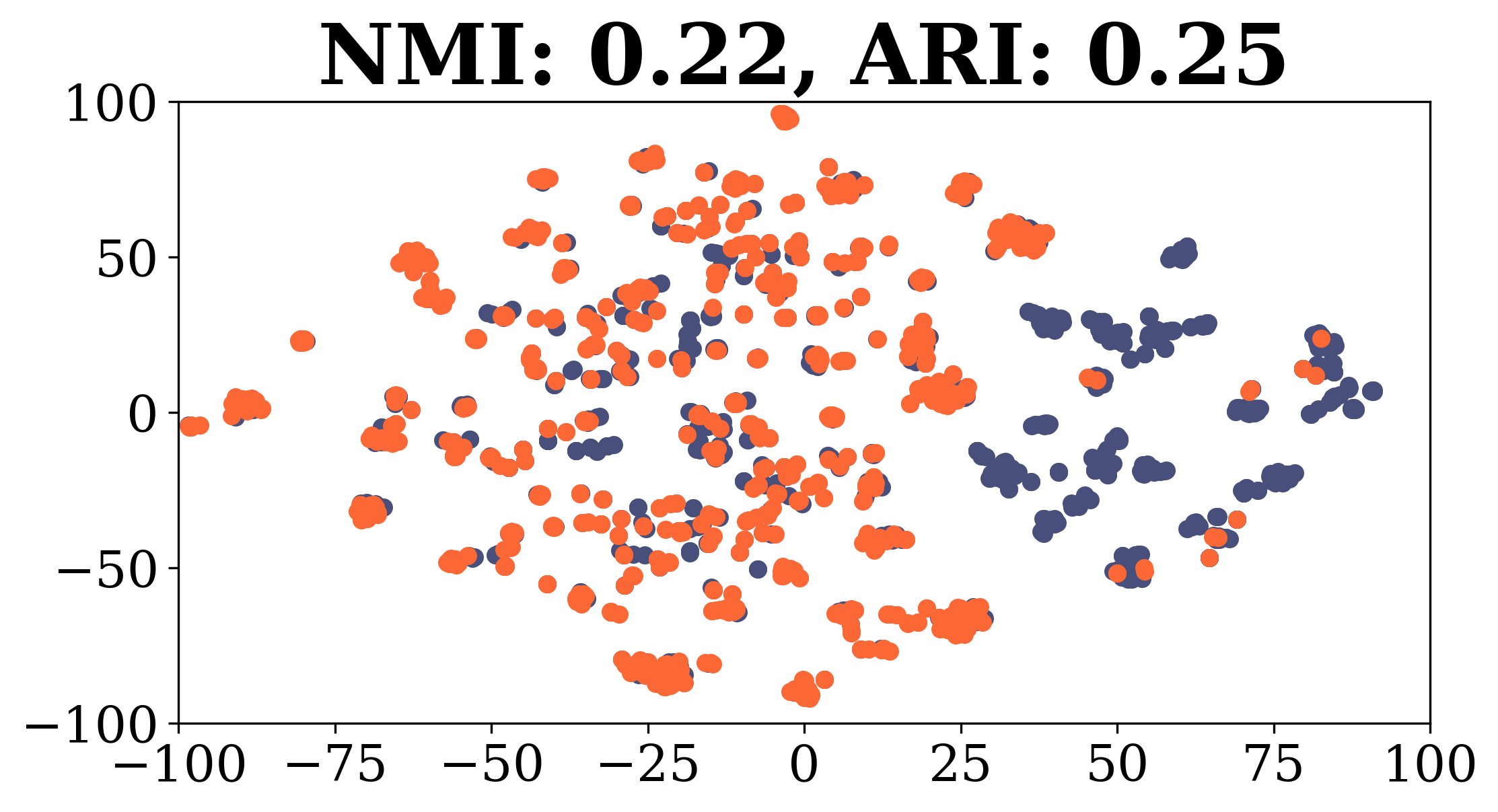}}
    \subfloat[NR-HGNN (Ours)]{\includegraphics[width=0.5\linewidth]{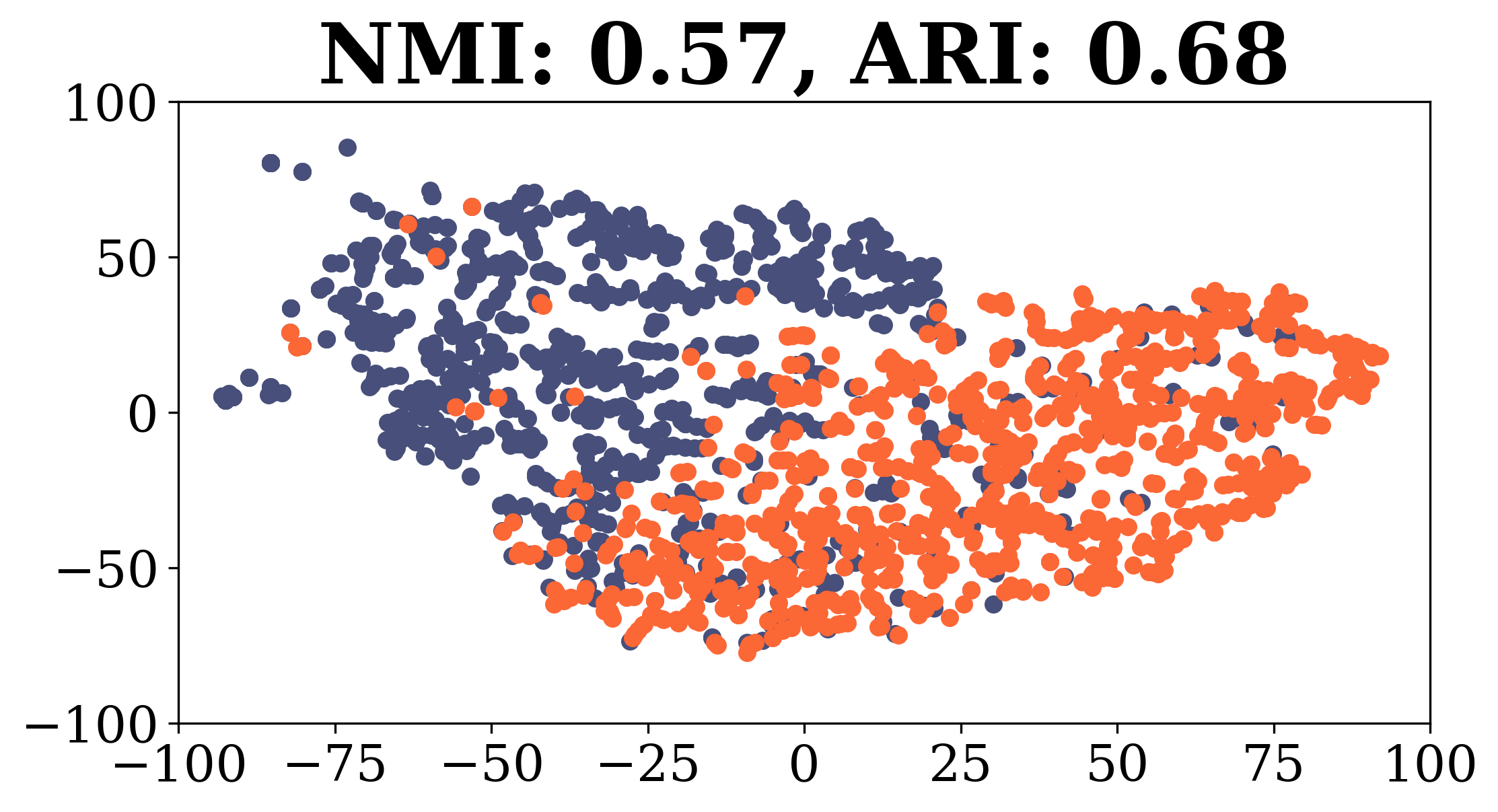}}
    \vspace{-10pt}
    \caption{T-SNE Visualization of opioid (blue) and non-opioid (orange) users embeddings.}
    \vspace{-10pt}
    \label{fig:embedding}
\end{figure}

\noindent\textbf{Visualization.} To elucidate the efficacy of our NR-HGNN model, we execute a visualization analysis using t-SNE, which elucidates the decision boundary between opioid and non-opioid users as depicted in Figure \ref{fig:embedding}. Relative to established methods such as GCN, RGCN, and the advanced SeHGNN, NR-HGNN exhibits superior clustering performance, as indicated by its compact intra-class distances and expansive inter-class distances. Conversely, the other three baselines tend to cluster users into smaller and more disparate groups. Furthermore, NR-HGNN demonstrates significantly higher Normalized Mutual Information (NMI) and Adjusted Rand Index (ARI) values, underlining its superiority. This enhanced performance may be attributed to NR-HGNN's ability to effectively discern reliable patterns within the \textit{NHANES Dietary Graph}.

\section{Deep Dive into the Dietary Patterns}

In addition to detecting potential opioid misuses, our Diet-ODIN also provides the provision of interpretive analyses of predictions. To this end, we select a random subset of 250 users, prompted with the dietary patterns with attention and users with shared dietary patterns using NR-HGNN. In alignment with the methodology delineated in Section 4.2, we conduct textual analysis and subsequent evaluations on the output. To vividly demonstrate our findings and the process, we deploy \textbf{a demonstration for our system}. One of the primary sections of our system is the statistical analysis panel to manifest the result of our qualitative analysis. We will dive deep into opioid users' dietary patterns by demonstrating the statistical analysis panel of Diet-ODIN. 

\subsection{Reasoning Analysis}

\begin{figure}[!t]
  \centering
  \vspace{-10pt}
  \includegraphics[width=\linewidth]{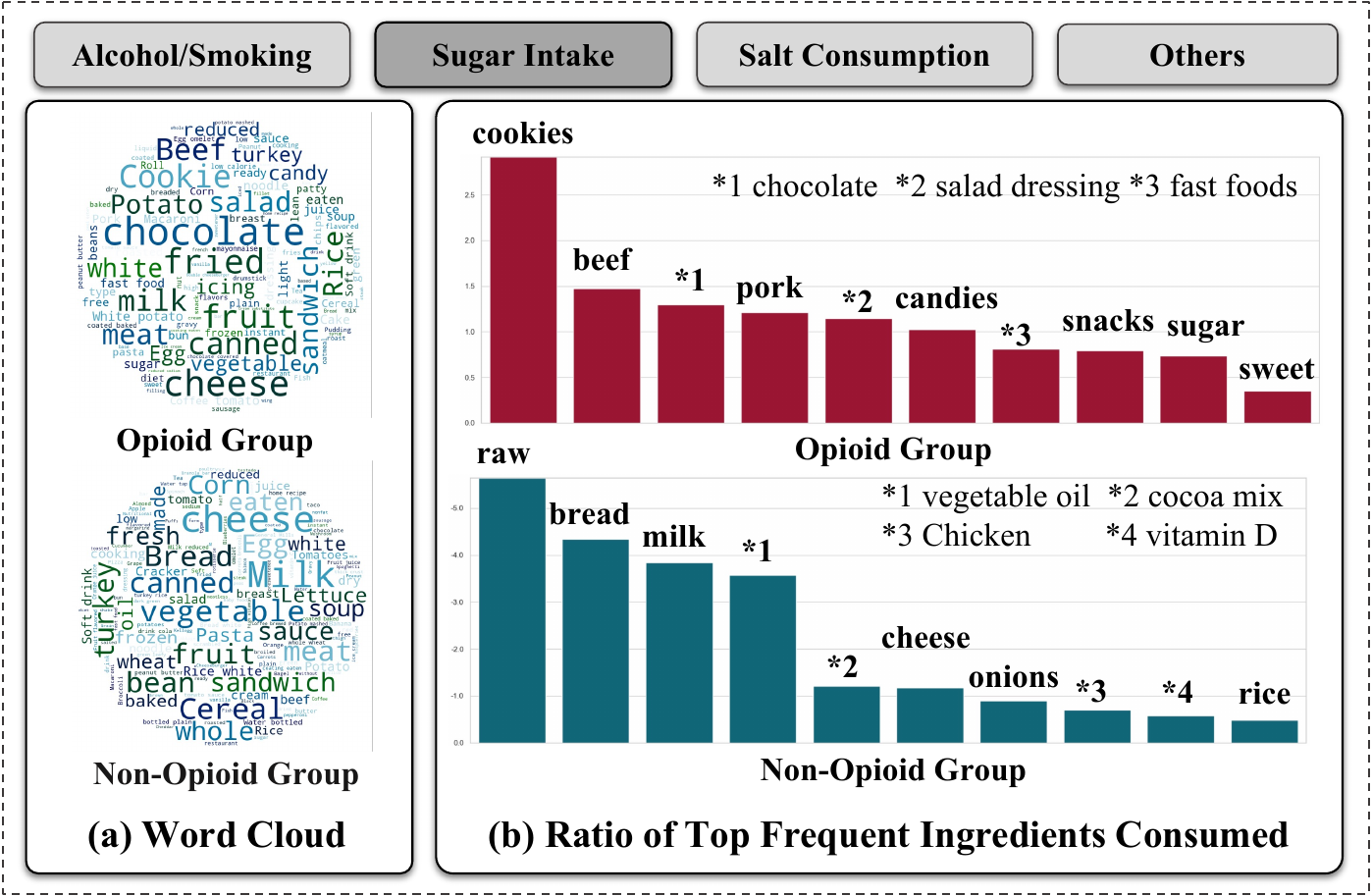}
  \vspace{-20pt}
  \caption{System Demo - Reasoning analysis on food patterns. (a) Word clouds generated from the ingredient names for both opioid and non-opioid groups.
  (b) The top ten frequent ingredients consumed by opioid and non-opioid groups. The analysis on dietary habits is in Appendix \ref{appendix:More Qualitative Analysis}.}
  \vspace{-5pt}
  \label{fig:foods}
\end{figure}

The statistical analysis panel categorizes and visually presents the results from the reasoning module. Take Sugar intake for example, Figure~\ref{fig:foods}.(a) demonstrates a word cloud contrasting keyword frequencies between opioid and non-opioid user groups. As we can see, the non-opioid group demonstrates a wider variety of dietary preferences, whereas the opioid group frequently references items like "chocolate", "cookie". This trend is echoed in the adjacent bar chart (Figure~\ref{fig:foods}.(b)), where the ingredient mentions of the opioid group, normalized based on group population size, reveal a similar pattern on sugar intensive ingredients. These results suggest a potential link between opioid use and sugar consumption. In addition to sugar intake, Diet-ODIN has identified several other noteworthy patterns. For example, dietary habits related to salt consumption, alcohol use, and smoking are prominent in our analysis, suggesting potential associations (as demonstrated in Appendix~\ref{appendix:More Qualitative Analysis}). To further identify the reliability of the conclusions drawn from the reasoning process, we provide analyses to show that the observations align with existing literature and statistical analysis in the section below.  


\subsection{Qualitative Analysis}

Diet-ODIN identifies key dietary factors—such as \textit{sugar intake}, \textit{salt consumption}, and \textit{alcohol and smoking behaviors}—as significant in detecting opioid misuse. To validate these findings, we analyze the NHANES dataset from 2003 onward, revealing patterns that support Diet-ODIN's conclusions. Specifically, a t-test comparing nutritional components between opioid users and non-users (Table~\ref{tab:nutrient_intake_analysis}) shows a notable difference in sugar and calorie intake, with a p-value less than 0.001, indicating statistical significance. This is further corroborated by related research \cite{mysels2010relationship, ochalek2021sucrose, avena2008evidence}. Moreover, a chi-square test is performed to examine dietary habits among opioid users (Table~\ref{tab:habit_analysis}), confirming a strong connection between salt consumption and opioid misuse. This association is reinforced by dietary habits related to salt intake and the literature \cite{smith2019influence, smith2018salt}, as well as by analyzing daily sodium intake (Table~\ref{tab:nutrient_intake_analysis}). Additionally, the connection between opioid misuse and alcohol and smoking behaviors is affirmed by statistical analysis (Table~\ref{tab:habit_analysis}) and literature review \cite{zvolensky2021opioid, ekholm2009alcohol}. Collectively, these analyses validate Diet-ODIN's effectiveness and reliability.

\begin{table}[!ht]
    \caption{
    Nutrient correlation analysis (t-test) reveals a strong correlation between potential opioid users and nutrient intake, corroborating Diet-ODIN's findings on sugar intake. }
    \vspace{-10pt}
    \centering
    \resizebox{0.85\linewidth}{!}{
        \begin{tabular}{lccc}
            \toprule
            \textbf{Nutrient} & \textbf{\makecell{Daily Avg.                          \\(Opioid Users)}} & \textbf{\makecell{Daily Avg. \\ (Regular Users)}} & \textbf{p-value} \\
            \midrule\midrule
            Sugar             & 111 (g)                      & 99 (g)      & $<0.001$ \\
            Calories          & 1867 (kcal)                  & 1726 (kcal) & $<0.001$ \\
            Sodium            & 2929 (mg)                    & 2749 (mg)   & $<0.001$ \\
            Caffeine          & 172 (mg)                     & 78 (mg)     & $<0.001$ \\
            \bottomrule
        \end{tabular}
    }
    \vspace{-10pt}
    \label{tab:nutrient_intake_analysis}
\end{table}

\begin{table}[!ht]
    \caption{
    Habit correlation analysis (chi-square test) reveals a tight linkage between certain dietary habits and potential opioid misuse, thereby validating Diet-ODIN's findings on alcohol and smoking behaviors, salt consumption, and new insights not previously explored.}
    \vspace{-10pt}
    \centering
    \setlength{\tabcolsep}{3pt} 
    \resizebox{\linewidth}{!}{
        \begin{tabular}{llccc}
            \toprule
             & \textbf{Habit Category}               & \textbf{\# \makecell{Opioid                    \\ Users}} & \textbf{\# \makecell{Total            \\ Users}} & \textbf{p-value} \\
            \midrule\midrule
             & NHANES Data                           & 2728                        & 95872 & -        \\
            \midrule\midrule
            \multirow{6}{*}{\rotatebox{90}{\makecell{\bf Alco. \& Smok.}}}
             & Uses tobacco rarely                   & 82                          & 1852  & $<0.001$ \\
             & Uses tobacco often                    & 383                         & 3155  & $<0.001$ \\
             & Drinks alcohol less than avg.      & 778                         & 26419 & $0.2627$ \\
             & Drinks alcohol more than avg.      & 835                         & 8634  & $<0.001$ \\
             & Light cigarette smoker                & 465                         & 11317 & $<0.001$ \\
             & Heavy cigarette smoker                & 164                         & 3398  & $<0.001$ \\
            \midrule\midrule
            \multirow{4}{*}{\rotatebox{90}{\bf Salt Cons.}}
             & Uses little to no salt in preparation & 764                         & 20596 & $<0.001$ \\
             & Uses lots of salt in preparation      & 912                         & 34341 & $<0.010$ \\
             & Adds little to no salt at table       & 674                         & 26604 & $<0.001$ \\
             & Adds lots of salt at table            & 499                         & 9654  & $<0.001$ \\
            \midrule\midrule
            \multirow{4}{*}{\rotatebox{90}{\bf New}}
             & Drinks little or no milk              & 503                         & 11717 & $<0.001$ \\
             & Drinks lots of milk                   & 1029                        & 45901 & $<0.001$ \\
             & Takes few or no supplements           & 1161                        & 53782 & $<0.001$ \\
             & Takes lots of supplements             & 770                         & 14027 & $<0.001$ \\
            \bottomrule
        \end{tabular}
    }
    \vspace{-10pt}
    \label{tab:habit_analysis}
\end{table}

\vspace{0.5cm}
\subsection{Dietary Pattern Discovery}    

Beyond identifying established dietary patterns related to opioid misuse, Diet-ODIN also unveils significant yet previously unexplored factors critical for detecting potential opioid misuse, providing new insights for future research aimed at combating opioid misuse. Notably, Diet-ODIN highlights the unique influence of \textit{milk consumption} and \textit{nutritional supplement intake} on opioid misuse, revealing that users with opioid misuse are (1) less likely to consume milk and (2) more inclined to rely on nutritional supplements instead of traditional food sources for their nutrient intake. This finding is further supported by statistical analysis (Table~\ref{tab:habit_analysis}) of the NHANES database. Such observations imply that users with opioid misuse may neglect their diet, however to our best knowledge, little literature has explored these specific directions. These emerging patterns warrant deeper investigation for future research avenues.

\section{Conclusion}


In this study, we introduce Diet-ODIN, an innovative framework designed to detect users with opioid misuse through interpretable analysis of dietary patterns. Diet-ODIN establishes a large-scale multifaceted dietary benchmark dataset and proposes the NR-HGNN model, which adeptly utilizes both shared and individual dietary patterns with a denoising refinement process. To uncover critical factors in opioid misuse identification, Diet-ODIN employs LLMs for reasoning and implements two innovative strategies to boost reasoning capabilities. The performance of NR-HGNN in detecting users with opioid misuse significantly exceeds that of various baseline models. Furthermore, our qualitative analysis verifies that the reasoning outcomes align with existing literature and statistical analysis. Moreover, we build a system to demonstrate these analyses, offering fresh perspectives, illuminating potential, more effective intervention strategies for addressing the opioid crisis.




\bibliographystyle{ACM-Reference-Format}
\bibliography{reference,ref-2}

\appendix
\section{Graph Construction Details}
\label{appendix:Graph Construction Details}

\paragraph{\textbf{Nutritional Information}}
The food records in NHANES data are recorded using Food and Nutrient Database for Dietary Studies (FNDDS) food code. This database, integral to USDA, catalogs food and beverage consumption in What We Eat In America (WWEIA) database and NHANES data to help researchers conduct enhanced analysis of nutrient values in dietary intakes. By leveraging this dataset, we link the food items in NHANES dataset with food ingredients and WWEIA food category information, and create \textit{food}, \textit{ingredient} and \textit{category} nodes.

\paragraph{\textbf{Dietary Habit Information}}
We compile dietary habit data from various NHANES tables, such as Diet Behavior table or Consumer Behavior table. Given the diverse nature of these features, traditional processing methods were inadequate. A team of four reviewers meticulously examines the NHANES data, identifying features indicative of dietary habits, such as the awareness towards healthy diet or the frequency of having frozen food. For each feature, we take the top 10\% and bottom 10\% of the users and assign these users the habit tags. For example, we take all those respond to milk drinking questionnaire, and assign the 10\% users who drink the most milk the habit “drink lots of milk”, and the 10\% users who drink the least the habit “drink little or no milk”. In this way, we summarize 54 distinct habits that serves as the \textit{habit} node in the graph. This habit information is of vital importance and faithfully reflect many insights into users' dietary patterns as we demonstrate later in the paper.

As for the node features, we use the demographic information as the features for the user nodes, and the nutrient values as the features for the food nodes. To associate other node types with their content, we use the pre-trained BERT embeddings to encode the description of the nodes, serving as the corresponding node representations. The edges record the connections between the nodes and are free of weights or directions. 

For the label for our benchmark dataset, we mark a user as an active user with opioid misuse if: 1) a user has records of using heroin within the year, or 2) a user has records of using a prescription opioid medicine continuously over 90 days, which is a criteria commonly employed in the domain~\cite{gu2022prevalence}. In this way, we generate the labels and form our problem into a binary classification task. The detailed graph statistics can be seen in Table~\ref{tab:graph_statistics}. 

\begin{table}[!h]
\caption{The statistics of NHANES Dietary Graph}
\vspace{-10pt}
\centering
\resizebox{\linewidth}{!}{
    \begin{tabular}{lc}
        \toprule
        \textbf{Item} & \textbf{Count}  \\ 
        \midrule
        \texttt{Nodes} &  \begin{minipage}{6.5cm}
            \# \textit{User} = 4,826, \# \textit{Food} = 5,896, \# \textit{Ingredient} = 2,792, \\
            \# \textit{Category} = 174, \# \textit{Habit} = 54
        \end{minipage}                       \\
        \midrule
        \texttt{Relations} &  \begin{minipage}{6.5cm}
            \# \textit{Food - User} = 136,967, \# \textit{Food - Ingredient} = 19,410, \\
            \# \textit{Food - Category} = 5,896, \# \textit{User - Habit} = 46,947
        \end{minipage}                       \\ 
        \bottomrule
    \end{tabular}
}
\label{tab:graph_statistics}
\vspace{-15pt}
\end{table}

\section{Experimental Settings}

\subsection{Environmental Settings}

The experiments are conducted in the Windows 10 operating system with 64GB of RAM. The training process involves the use of one NVIDIA GeForce RTX 3090 GPU and one NVIDIA A40 GPU, with the framework of Python 3.8.18, Pytorch 2.1.0 and Pytorch-geometric 2.4.0.

\subsection{Split Settings}

We randomly split the data into a 2-4-4 group (20\% training set, 40\% valid set and 40\% test set), and 4-3-3 group and conduct experiments on these two split settings. The random split is only performed once and all experiments are done on the same split seed. 

For the noise reduction phase, since we use the user label to identify whether two users share the same label, we filter the user pairs to make sure the two users always come from the training set to avoid data leak. Also, at this stage, we use the model from the last training epoch instead of the one performs best on the validation set to avoid data leakage on valid data during inference.

\subsection{Hyperparameters}
In our experiments, for the macro-aggregation stage, we set the hidden dimension to 64, and in micro-aggregation stage to 256. Both stages involve 4 attention heads. The pooling layer afterwards use a simple MLP with the concatenated hidden dimension of the two model as the input, and has hidden dimension of 256 as output. We perform Adaptive Moment Estimation (Adam) to optimize the models with a learning rate of 0.001 and L2 regularization as 0.001 for 100 epochs. For the subsequent opioid user prediction module, we train a simple GCN layer for 500 epochs, which has a hidden dimension of 128, a learning rate of 0.0001, a L2 regularization as 0.001 and a dropout rate of 0.6. All experiments utilize the cross entropy as the loss function. In the training, all models are evaluated on the valid sets and the model state with the best valid set performance is used to evaluate the test set. To improve reproducibility, for each split setting, we evaluate the models using the same set of 10 consecutive seeds, and report their mean and standard deviation. 

\section{Fine-grained Correlation between Dietary Patterns and Opioid Misuse}

To further analyze the fine-grained factors in detecting users with opioid misuse, we would analyze the impact of the ingredients to the foods, enabling us to discover more intricate patterns. In particular, we also utilize the attention score extracted from Contextual Aggregation (Section \ref{sec:con aggr}) to rank the importance of ingredients in detecting opioid users. We employ a similar context prompt, which is formulated as ``\textit{Food A is <food description>. The model thinks the most important ingredients are: ingredients 1: <>, Ingredient 2: <>, ..., and the remaining ingredients 1: <>, Ingredient 2: <>, ..., and }''. Through the more fine-grained ingredient level analysis, this process helps us navigate to the most important ingredients that contributes to the identification of users with opioid misuse. For example, when cakes with chocolate icing is identified, we can further see it's the ingredient chocolate, instead of eggs or flour that makes the food indicative.

\section{More Qualitative Analysis}
\label{appendix:More Qualitative Analysis}
In this section, we further elaborate on the reasoning analysis and the qualitative analysis we do that demonstrate Diet-ODIN's efficacy and reliability.

\begin{figure}[!ht]
  \centering
  \includegraphics[width=\linewidth]{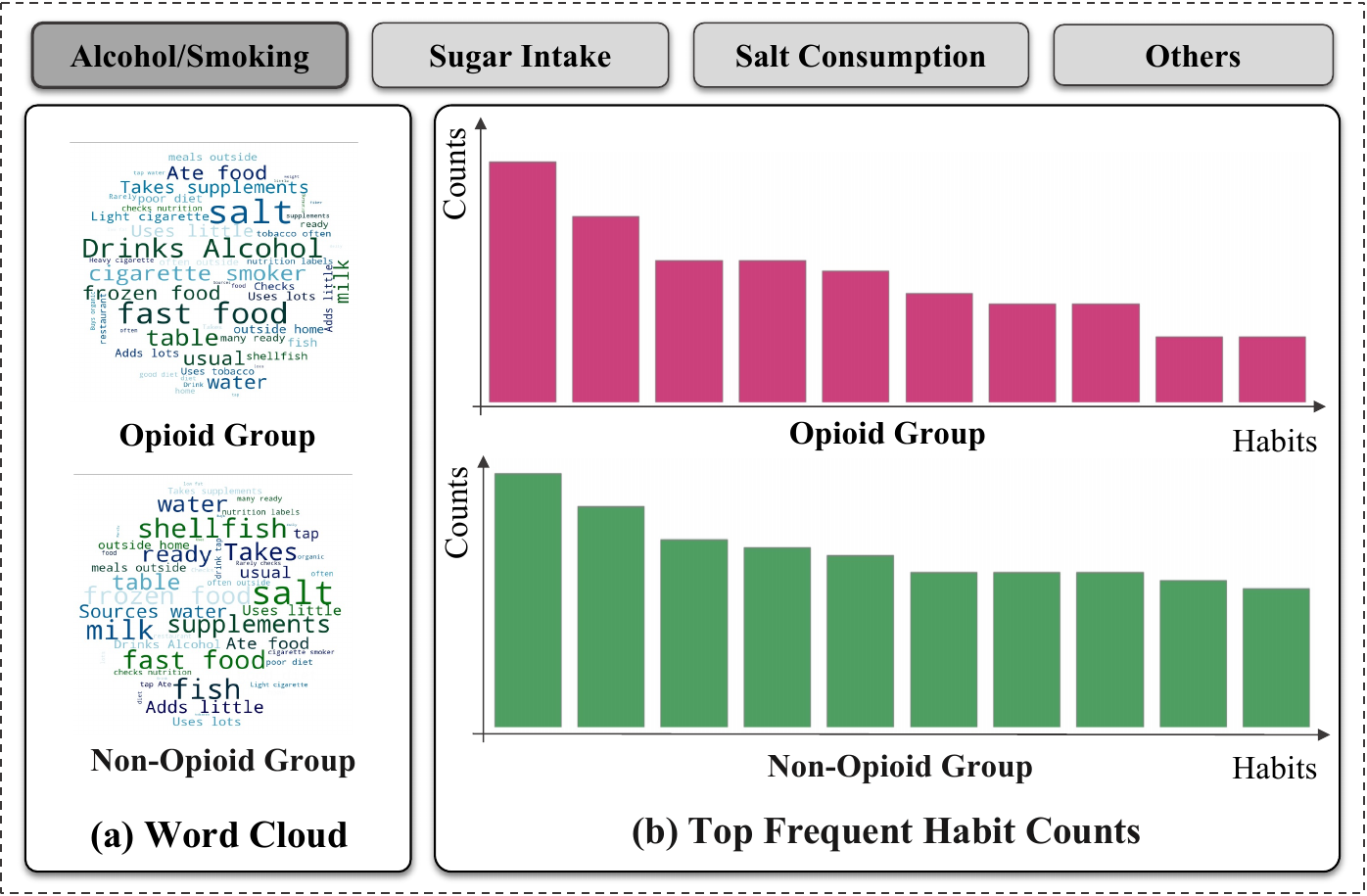}
  \vspace{-20pt}
  \caption{Reasoning analysis for habits: (a) Word cloud generated from the habit names for both opioid and non-opioid groups. (b) The top ten frequent habit counts by opioid and non-opioid groups. Because of the length of the habit names, we hide the name captions and refer by its column sequence. For example, $op1$ and $nonop-1$ refers to the leftmost bar of upper and lower barplot respectively. }
  \label{fig:habits}
\end{figure}

Figure~\ref{fig:habits} illustrates the analysis of alcohol and smoking habits, as selected via the top buttons in the system. The bar chart below the buttons presents the most frequent habits counts among opioid and non-opioid user groups from the reasoning result. The chart includes four specific habits related to alcohol and cigarette use (e.g., $op1$-drinking alcohol more, $nonop6$-drinking alcohol less than average, $op2$-heavy cigarette smoking, $nonop9$-light cigarette smoking), indicating a significant correlation between opioid usage and alcohol or cigarette consumption. Additionally, a word cloud, positioned next to the bar chart, highlights the most common terms associated with these habits. Notably, terms like "drinks alcohol" and "cigarette smoker" predominate in the opioid user group. Similar patterns can be found in salt consumption. Opioid users tend to have habits related to using more salt, such as $op4$-Adds lots of salt at table, and $op6$-Uses lots of salt in preparation. 
Similarly, in next paragraphs, we utilize statistical analysis and existing literature to show Diet-ODIN's reasoning is aligned with real-world scenarios.

\textbf{Opioid Use and Salt Consumption.}
As evidenced by Table~\ref{tab:habit_analysis}, all patterns related to salt consumption are statistically correlated. This finding is further substantiated by the analysis of daily average sodium intake in Table~\ref{tab:nutrient_intake_analysis}, where the p-value indicates a statistically significant difference between the opioid user group and the non-opioid user group. The literature corroborates this pattern as well. For example, as Smith et al~\cite{smith2019influence, smith2018salt} discussed in their research that opioid dependence can affect dietary salt intake, often resulting in a preference for foods with higher salt concentrations.

\textbf{Opioid Use and the Use of Alcohol and Smoking.}
In a manner akin to the examination of salt consumption, numerous studies have investigated the relationship between opioid usage and the consumption of other substances. A well-established body of research indicates a positive correlation between opioid use and excessive alcohol consumption and smoking behavior, as noted in studies by \cite{zvolensky2021opioid, ekholm2009alcohol}. These associations are also detected in our analysis using the NHANES data. The majority of the alcohol and cigarette use patterns identified by our method have been corroborated as correlational. (We amalgamated 'tobacco use' and 'cigarette smoking' during the reasoning phase.)

\vspace{-2pt}

\section{Limitations and Discussion}
Although our developed \textit{Diet-ODIN} has been demonstrated its outstanding performance on exploring  the complex interplay between opioid misuse and dietary patterns, it could also be subject to certain limitations: data incompleteness and lacking quantitative measure of the interpretations. For data incompleteness, in this work, the quantities of foods are not accessible where analysis bias may exist. Due to the lack of food quantities, the developed model is not able to incorporate  edge weights to capture more detailed dietary patterns, which is inclined to answer how frequent a user eats a certain food. Nevertheless, edge weights might be helpful in certain scenarios, for example, our method doesn't consider ``coffee'' an important factor because it is such a common drinking in daily life that everybody would drink. However, the drinking amounts weigh differently in one's dietary structure. In table-\ref{tab:nutrient_intake_analysis}, we demonstrate that opioid users on average consume more than double amount of caffeine than that of the regular users, which should be an indicator captured by both detection and reasoning tasks. This limitation can be mitigated by implementing edge-weighting scheme on user-food edges. Nevertheless, our experiments and analysis demonstrate the majority of dietary patterns can be captured through edge-weight free methods. Thus, we leave the edge weighting implementation for our future work. For the lack of quantitative measure of the interpretations, our reasoning experiments mainly rely on statistical qualitative analysis, which posts influence by reflecting real-life patterns and pointing out future research directions. The consistent matching findings between Diet-ODIN and real-life statistics and literature indicate the efficacy and reliability of the patterns. However, the limitation remains that the reasoning process lacks control and quantitative benchmarks to evaluate its efficacy. We argue that our work set a strong baseline by providing insights and setup benchmarks that connect dietary and healthcare information in NHANES data. Despite these limitations, our work in this paper is the first attempt to exploit AI-driven techniques for comprehensive exploration of the correlations between opioid misuse and users' diets with explainable outputs, which could provide new insights and pave promising research directions for other researchers and practitioners to combat the devastating and lethal opioid epidemic.

\end{document}